\documentclass{article}
\usepackage{ijcai23}

\usepackage{multirow}
\usepackage{pifont}
\usepackage{amssymb}
\usepackage{marvosym}

\usepackage{times}
\usepackage{soul}
\usepackage{url}
\usepackage[hidelinks]{hyperref}
\usepackage[utf8]{inputenc}
\usepackage[small]{caption}
\usepackage{graphicx}
\usepackage{amsmath}
\usepackage{amsthm}
\usepackage{booktabs}
\usepackage{algorithm}
\usepackage{algorithmic}
\usepackage[switch]{lineno}

\title{Timestamp-Supervised Action Segmentation from the Perspective of Clustering}

\author{
Dazhao Du$^{1,2}$
\and
Enhan Li$^{2,3}$\and
Lingyu Si$^{1,2}$\and
Fanjiang Xu$^1$\textsuperscript{(\Letter)}\And
Fuchun Sun$^{1,4}$
\affiliations
$^1$Institute of Software, Chinese Academy of Science\\
$^2$University of Chinese Academy of Sciences\\
$^3$Institute of Computing Technology, Chinese Academy of Sciences\\
$^4$Department of Computer Science and Technology, Tsinghua University
\emails
dudazhao20@mails.ucas.ac.cn,
lienhan20g@ict.ac.cn,
\{lingyu,fanjiang\}@iscas.ac.cn,
fcsun@mail.tsinghua.edu.cn
}

\begin{document}

\maketitle

\begin{abstract}
    Video action segmentation under timestamp supervision has recently received much attention due to lower annotation costs. Most existing methods generate pseudo-labels for all frames in each video to train the segmentation model. However, these methods suffer from incorrect pseudo-labels, especially for the semantically unclear frames in the transition region between two consecutive actions, which we call \textbf{ambiguous intervals}. To address this issue, we propose a novel framework from the perspective of clustering, which includes the following two parts. First, \textbf{pseudo-label ensembling} generates incomplete but high-quality pseudo-label sequences, where the frames in ambiguous intervals have no pseudo-labels. Second, \textbf{iterative clustering} iteratively propagates the pseudo-labels to the ambiguous intervals by clustering, and thus updates the pseudo-label sequences to train the model. We further introduce a clustering loss, which encourages the features of frames within the same action segment more compact. Extensive experiments show the effectiveness of our method.
\end{abstract}

\section{Introduction}

\begin{figure}[t]
\begin{center}
\includegraphics[width=0.8\columnwidth]{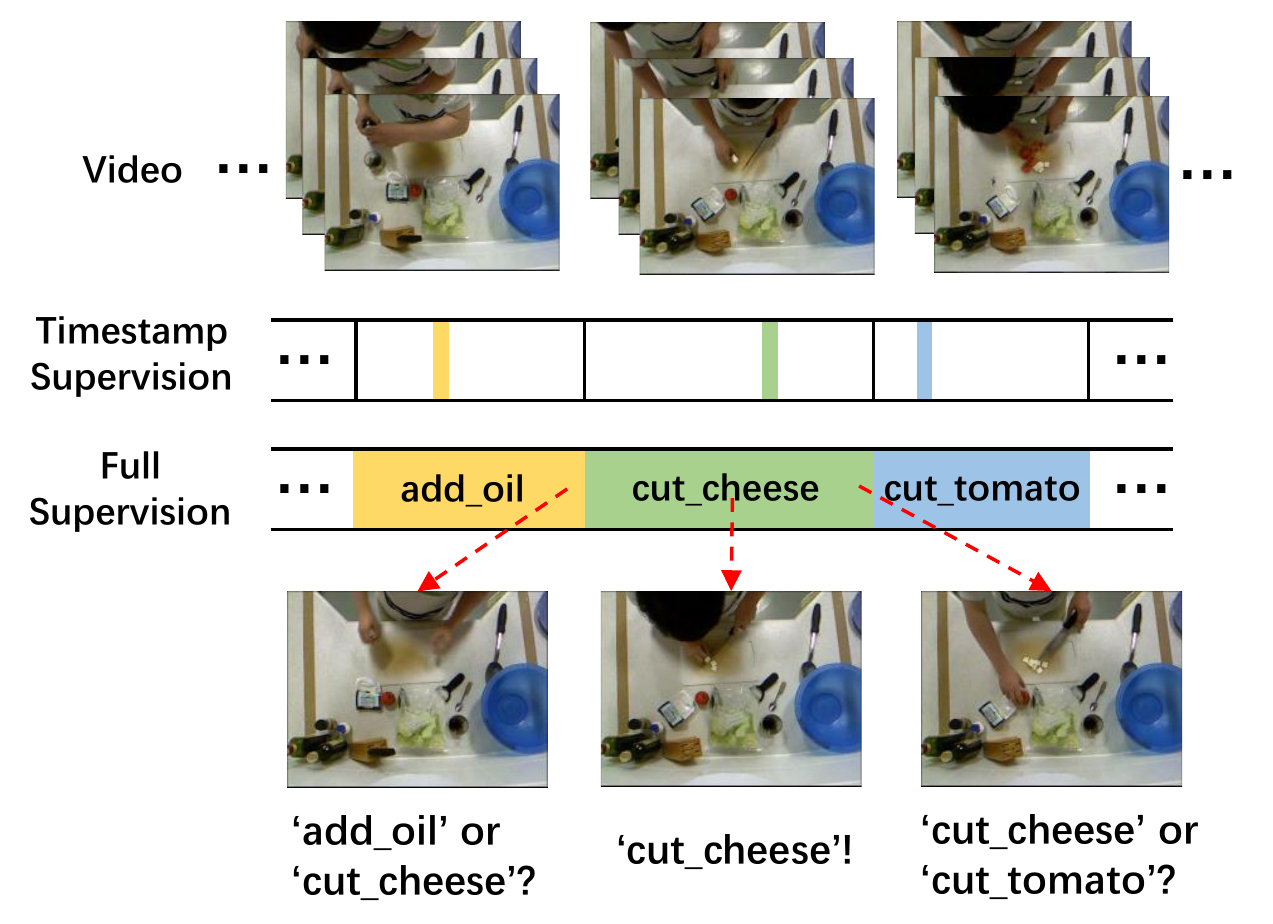}
\caption{Only one frame in each action segment is labeled under timestamp supervision, while all frames are labeled under full supervision. Those frames near the end of action ‘add\_oil’ can be classified as either action ‘add\_oil’ or 'cut\_cheese'. On the contrary, the frames near the middle of the action segment are usually representative of the action class.}
\label{fig:motivation}
\end{center}
\vspace{-0.4 cm}
\end{figure}

Recognizing and segmenting actions from long untrimmed videos is a challenging task. Action segmentation aims to classify actions for all frames in the video, resulting in several action segments. There have been many fully supervised approaches achieving good results. These approaches mainly use TCN~\cite{farha2019ms,li2020ms} or Transformer~\cite{TUT} to model the long-term dependencies between frames. However, obtaining label sequences for long videos is costly. Researchers have proposed several weaker levels of supervision, e.g., set supervision~\cite{actionset}, transcript supervision~\cite{kuehne2017weakly}, and timestamp supervision~\cite{li2021temporal}. Among them, timestamp supervision has the most promising performance.

\begin{figure}[t]
\begin{center}
\includegraphics[width=0.8\columnwidth]{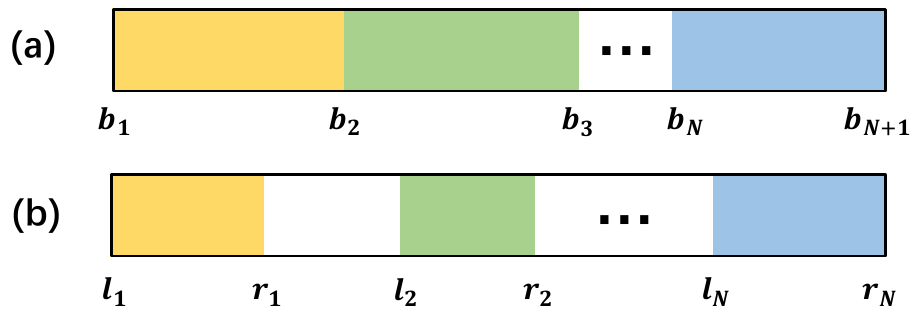}
\caption{Two types of pseudo-label sequences for a video containing $N$ action segments. (a) In the full pseudo-label sequence without ambiguous intervals, action segments are connected head-to-tail. $\{b_1,b_2,...,b_{N+1}\}$ contains the boundaries of all the actions, including the first frame and last frame of the video. (b) There exist ambiguous intervals between action segments. $l_n$ and $r_n$ are the left and right boundaries of the $n$-th action segment.}
\label{fig:notation}
\end{center}
\vspace{-0.3 cm}
\end{figure}

As shown in Figure~\ref{fig:motivation}, timestamp supervision only labels one frame in each action segment. Compared to other weakly supervised setups, timestamp supervision provides temporal information that is crucial for action segmentation and therefore has better performance. However, it is surprising that timestamped supervision sometimes outperforms the fully supervised setup. The reason is that the boundaries of action segments are usually ambiguous, which might result in inconsistencies between labels labeled by different annotators under full supervision. For example, the frames near the boundary between action ‘add\_oil’ and action ‘cut\_cheese’ in Figure~\ref{fig:motivation}, can be classified as either action ‘add\_oil’ or action ‘cut\_cheese’. We call the intervals close to the boundaries \textbf{ambiguous intervals}. In contrast, timestamp supervision usually labels the representative frame of an action, such as a frame near the middle of the action segment. The most popular method~\cite{li2021temporal} iteratively generates full pseudo-label sequences by detecting action changes, as shown in Figure~\ref{fig:notation} (a). However, this might lead to error accumulations, i.e. training with wrong pseudo-labels of ambiguous intervals will cause the model to deviate further from expectations. Therefore, we do not expect to generate pseudo-labels for frames in ambiguous intervals at the beginning, as shown in Figure~\ref{fig:notation} (b). As the training progresses, we can gradually assign pseudo-labels with high certainty to ambiguous intervals based on the model outputs.

We implement this idea from a clustering perspective by considering each action segment as a cluster. Our framework consists of two main parts, namely \textbf{pseudo-label ensembling (PLE)} and \textbf{iterative clustering (IC)}. First, PLE generates high-quality pseudo-label sequences with ambiguous intervals in Figure~\ref{fig:notation} (b). Specifically, we utilize three different clustering algorithms to generate three full pseudo-label sequences for each video. For each frame, if the three algorithms generate inconsistent pseudo-labels, the frame belongs to the ambiguous interval. Frames in the ambiguous intervals have no pseudo-labels and therefore do not affect training. Afterward, IC gradually assigns the frames in ambiguous intervals to left or right action segments by feature clustering based on their distance from the surrounding cluster centers. During the iterative training, the model can generate better features, based on which propagating pseudo-labels to the ambiguous interval results in more consistent and accurate supervision. As this process involves feature clustering, we further introduce the clustering loss commonly used in deep clustering to help feature learning and clustering.

In summary, our main contributions are twofold: (1) We design a framework for timestamp-supervised action from the perspective of clustering. The framework mainly contains two novel algorithms, \textbf{pseudo-label ensembling (PLE)} and \textbf{iterative clustering (IC)}, to address the negative impact of ambiguous intervals on training. Besides, we introduce the clustering loss. (2) Extensive experiments and ablation studies demonstrate the effectiveness of our framework and illustrate the impact of each component.

\section{Related Work}

\paragraph{Fully Supervised Action Segmentation.} In the fully supervised setting, all the frames are labeled during training. 
% Earlier approaches~\cite{rohrbach2012database,karaman2014fast} apply sliding windows and use non-maximum suppression to obtain the action segments. 
To model the temporal dynamics, some researchers utilize graphical models and sequential models such as HMM~\cite{tang2012learning,kuehne2016end}, RNN~\cite{singh2016multi}.
With the development of deep learning, temporal convolutional networks (TCNs) have been successful due to their powerful temporal modeling ability~\cite{lea2017tcn,farha2019ms,li2020ms,wang2020bcn,asrf}. 
% There are also some works focusing on combining boundary prediction and TCN-based models~\cite{wang2020bcn,asrf}. 
Recently, ASFormer~\cite{ASformer} and TUT~\cite{TUT} have designed transformer-based architectures that outperform TCN-based models. In contrast to these methods, our framework is model agnostic and relies on timestamp supervision instead of full supervision.

\paragraph{Weakly Supervised Action Segmentation.} Many works have attempted to explore weaker levels of supervision to reduce annotation costs. Transcript supervision provides ordered action lists. Some researchers introduce connectionist temporal classification~\cite{ECTC}, dynamic time warping~\cite{d3tw}, and energy-based learning~\cite{cdfl} to align the frame-wise outputs to transcripts. Other approaches utilize the Viterbi algorithm on the output probabilities to generate pseudo-label sequences~\cite{kuehne2017weakly,nnviterbi}. Another form of supervision, set supervision~\cite{actionset,sct,acv}, provides only the unordered action sets. However, these weakly supervised methods perform considerably worse than fully supervised approaches.

\paragraph{Timestamp-Supervised Action Segmentation.} 
% Since the concept of timestamp supervision was introduced to untrimmed video action recognition~\cite{moltisanti2019action}, some researchers have applied it to action segmentation. 
In action segmentation, Li et al.~\shortcite{li2021temporal} firstly proposed a timestamp-supervised framework that detected action changes by the energy function and generated full pseudo-label sequences to train the segmentation model. Based on it, Khan et al.~\shortcite{khan2022timestamp} used learnable GNN instead of the energy function to obtain action boundaries, achieving better performance. UVAST~\cite{uvast2022ECCV} extended the K-medoids clustering algorithm to yield pseudo-labels for all frames. Rahaman et al.~\shortcite{rahaman2022generalized} utilized the Expectation-Maximization algorithm to solve this problem. However, the loss function is still calculated on all frames to train the model during the M-step. Different from these methods, we iteratively generate pseudo-label sequences with ambiguous intervals, rather than generating pseudo-labels for all frames.

\paragraph{Weakly Supervised Learning.} When faced with the challenge of training samples lacking labels, one common approach is to assign pseudo-labels to all unlabeled samples and use them to train the model~\cite{lee2013pseudo}. However, we do not generate pseudo-labels for all unlabeled frames, considering the presence of ambiguous intervals. Iscen et al.~\shortcite{iscen2019label} constructed a similarity graph among samples based on features to propagate labels, while we propagate labels by iterative clustering. There are also some works learning with coarse-grained labels~\cite{qwwqwq} or noisy labels~\cite{qiang2021robust}.

\section{Methodology}

\subsection{Problem Definition}

Given an input long video which is represented by a sequence of frame-wise visual features: \({X} = [{x}_1, \cdots, {x}_T]\), where \({x}_t\) is the visual feature of the \(t\)-th frame. Our goal is to predict the action class label of each frame \([\hat{c}_1, \cdots, \hat{c}_T]\). Under timestamp supervision, only a single frame in each action segment is annotated. For a video having $N$ action segments, timestamp supervision provides a set of annotated timestamps $\{c_{\tau_1}, \cdots, c_{\tau_N}\}$, where the $\tau_n$-th frame is in the $n$-th segment.

\subsection{Overview}

\begin{figure*}[t]
\begin{center}
\includegraphics[width=0.8\linewidth]{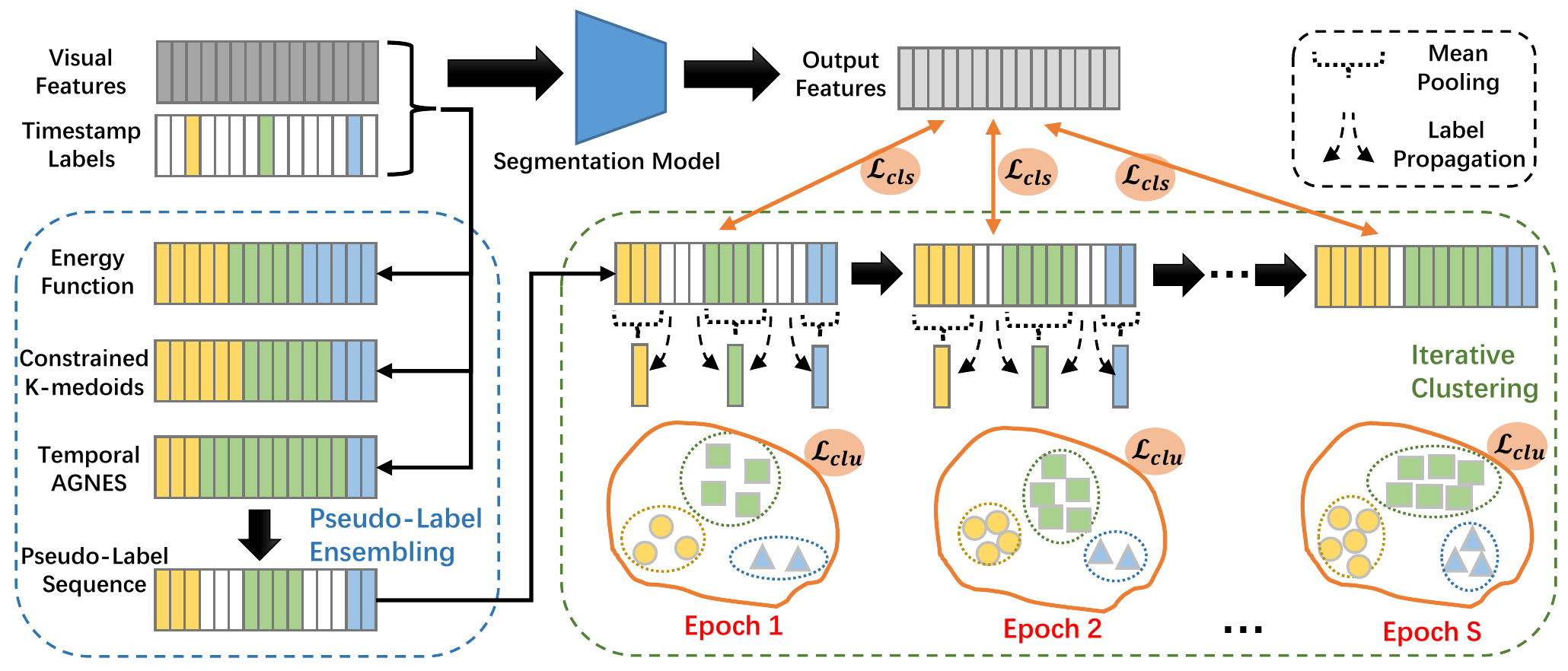}
\caption{The overall framework of our method. Given frame-wise visual features and timestamp supervision, pseudo-label ensembling can generate pseudo-label sequences with ambiguous intervals. Subsequently, in each epoch of iterative clustering, the pseudo-labels are propagated into ambiguous intervals by clustering, and the resulting new pseudo-label sequences will train the segmentation model. $\mathcal{L}_{cls}$ and $\mathcal{L}_{clu}$ are classification loss and clustering loss, respectively.}
\label{fig:overview}
\end{center}
\vspace{-0.3 cm}
\end{figure*}

As shown in Figure~\ref{fig:overview}, we design a novel framework for timestamp-supervised action segmentation, which contains pseudo-label ensembling (PLE) and iterative clustering (IC). Given the visual features and timestamp labels of each video, PLE combines three different clustering algorithms to generate pseudo-label sequences, where the frames in ambiguous intervals have no pseudo-labels while other frames have. We regard PLE as a pre-processing step since it relies on only the visual features of frames. During the first training epoch in IC, we utilize the pseudo-label sequences generated by PLE to train the segmentation model. After training, we cluster the output features of the model. Specifically, we consider frames with the same pseudo-label to be in the same cluster, which means that an action segment is a cluster. The center vector of each cluster can be calculated based on the output features of all frames within the action segment. We classify the frames within ambiguous intervals according to the distances between their output features and surrounding cluster centers, which can propagate pseudo-labels and narrow down the ambiguous intervals. During the next training epoch, the model is trained by the updated pseudo-label sequences. We iterate this process for $S$ epochs. The additional clustering loss gathers features of the same class. We describe PLE, IC, and the clustering loss in detail in the following subsections.

\subsection{Pseudo-Label Ensembling}

We find that action segmentation under timestamp supervision can be viewed as a special clustering problem. For a video containing $N$ action segments, we can cluster the features of frames into $N$ clusters. Since each timestamp will belong to a certain cluster, we label all frames in that cluster with the action label of the timestamp. In PLE, we ensemble three different clustering algorithms, including energy function~\cite{li2021temporal}, constrained k-medoids~\cite{uvast2022ECCV}, and temporal AGNES we propose. 

\paragraph{Energy Function.} Given frame-wise visual features and labeled timestamps in a video, the energy function aims to find action changes $\{b_2,b_3,...,b_N\}$ by minimizing the variations of the features within each of two consecutive segments, as shown in Figure~\ref{fig:notation} (a).

\paragraph{Constrained K-medoids.} Similar to the energy function, the constrained k-medoids algorithm also aims to find the temporal boundaries of each segment such that the accumulative distance of each cluster to the current cluster center is minimized. Different from the typical k-medoids algorithm, it forces the clusters to be temporally continuous.

\paragraph{Temporal AGNES.} AGNES(Agglomerative Nesting)~\cite{agnes} is a bottom-up hierarchical clustering algorithm: each frame starts in its own cluster, and clusters are successively merged together. We propose temporal AGNES, which only merges temporally adjacent clusters. The clustering algorithm stops when the number of clusters equals the number of timestamps. The detailed steps of the temporal AGNES algorithm are shown in Appendix~\ref{sec:agnes}.

\paragraph{PLE.} For each video sample, we can generate three different full pseudo-label sequences by the above algorithms. The frames within ambiguous intervals are semantically ambiguous, and different clustering algorithms might assign different action labels to these frames. Inspired by ensembling learning~\cite{ensemble}, we take the intersection of these three pseudo-label sequences. If the three algorithms assign the same pseudo-label to a frame, then that frame is labeled as this action class; otherwise, the frame belongs to the ambiguous interval and has no pseudo-label. So we can obtain a pseudo-label sequence with ambiguous intervals, as shown in Figure~\ref{fig:notation} (b). The pseudo-label sequence generated by PLE achieves a trade-off between label rate and accuracy. Compared to the timestamp labels, it has a higher label rate. Compared to the full pseudo-label sequence generated by any clustering algorithm, it has a higher accuracy. More analysis can be found in Figure~\ref{fig:ple} and Table~\ref{tab:abla_ple}.

\subsection{Iterative Clustering}

From the perspective of clustering, IC iteratively generates pseudo-labels for frames within ambiguous intervals to update the pseudo-label sequences for training the segmentation model $\mathcal{M}$. We clarify some notations before introducing IC in detail. $x_t$ and $h_t$ are the visual feature and output feature of the $t$-th frame, respectively. As shown in Figure~\ref{fig:notation}, $l_n$ and $r_n$ are the left and right boundaries of the $n$-th action segment in the pseudo-label sequence. $m_n$ is the mean vector of all the output features in the $n$-th action segment, which is equivalent to the cluster center vector of the $n$-th cluster. $S$ and $N$ are the numbers of epochs and timestamp labels, respectively. Function $\textup{d}(\cdot,\cdot)$ calculates the Euclidean distance between the output features of two frames.

\begin{algorithm}[t]
    \small
    \caption{The iterative clustering algorithm}
    \label{alg:IC}
    \renewcommand{\algorithmicrequire}{\textbf{Input:}}
    \renewcommand{\algorithmicensure}{\textbf{Output:}}
    \begin{algorithmic}[1]
        \REQUIRE visual features: $[x_1,x_2,...,x_T]$, pseudo-label sequence: $[\hat{c}_1,\hat{c}_2,...,\hat{c}_T]$, timestamp supervision: $\{c_{\tau_1},c_{\tau_2},...,c_{\tau_N}\}$, initial segmentation model: $\mathcal{M}$
        \ENSURE segmentation model after iterative training: $\mathcal{M}$
        \STATE $[\widetilde{c}_{1},\widetilde{c}_{2},...,\widetilde{c}_{T}] \gets [\hat{c}_1,\hat{c}_2,...,\hat{c}_T]$ 

        % \COMMENT{Initialize pseudo-label sequence}
        \STATE Obtain the sets of left and right action segment boundaries from the pseudo-label sequence: $\{l_1,l_2,...,l_N\}$ and $\{r_1,r_2,...,r_N\}$
        
        \FOR{epoch $=1,2,...,S$}
            \STATE $[h_1,...,h_T] \gets \mathcal{M}([x_1,...,x_T])$ 
            % \COMMENT{Forward propagation}
            
            \FOR{$n=1,2,...,N$}
                \STATE $m_n \gets \frac{1}{r_n-l_n}\sum_{t=l_n}^{r_n}h_t$ 
                % \COMMENT{Calculate the centroid vector of the action cluster}
            \ENDFOR
            
            \FOR{$n=1,2,...,N-1$}
            % \COMMENT{Label propagation for the ambiguous intervals}
                \WHILE{$r_n<l_{n+1}$ and $\textup{d}(h_{r_n},m_n)<\textup{d}(h_{r_n},m_{n+1})$}
                \STATE $r_n \gets r_n+1$ 
                % \COMMENT{Update the right border of the left action segment}
                \ENDWHILE
                
                \WHILE{$r_n \leq l_{n+1}$ and $\textup{d}(h_{l_{n+1}},m_n) > \textup{d}(h_{l_{n+1}},m_{n+1})$}
                \STATE $l_{n+1} \gets l_{n+1}-1$ 
                % \COMMENT{Update the left border of the right action segment}
                \ENDWHILE
                \STATE $l_{n+1} \gets l_{n+1}+1$ 
                
            \ENDFOR
            
            \FOR{$n=1,2,...,N$}
            % \COMMENT{Update the pseudo-label sequence}
                \STATE $[\widetilde{c}_{l_n},\widetilde{c}_{l_n+1},...,\widetilde{c}_{r_n-1}] \gets [c_{\tau_n},c_{\tau_n},...,c_{\tau_n}]$ 
                % \COMMENT{Belong to the same action segment}
            \ENDFOR

            \STATE Train the model for one epoch with the pseudo-label sequence $[\widetilde{c}_1,...,\widetilde{c}_T]$, and obtain the new segmentation model $\mathcal{M}$ after parameter update
            
        \ENDFOR
        
    \end{algorithmic}
\end{algorithm}

The algorithm~\ref{alg:IC} summarizes the steps of our iterative clustering algorithm. Steps 4 to 20 describe the procedure for each epoch. First, the visual features of the video frames are fed into the segmentation model to generate output features (step 4). The center vector of each action cluster is calculated based on output features and the pseudo-label sequences from the previous epoch (steps 5-7). Afterward, for every ambiguous interval, we first iterate through each frame in the ambiguous interval from left to right and compare the distance between that frame and the center vectors of the adjacent left and right action clusters (steps 9-11). If it is closer to the center vector of the left cluster, the frame belongs to the left action segment. We stop the iteration when the frame is closer to the center vector of the right cluster, which means that the frame either belongs to the right action segment or remains in the ambiguous interval. This iterative process is pseudo-label propagation, which propagates the pseudo-labels from left to right to the frames in the ambiguous interval. Subsequently, we propagate the pseudo-labels from right to left (steps 12-15), which is symmetric to the above process. After finishing pseudo-label propagation, the range of ambiguous intervals is reduced and the resulting supervision signal is more intensive. Finally, the segmentation model is trained by the new pseudo-label sequence (step 20).

\subsection{Loss Function}
During the training phase, we combine four different losses.

\paragraph{Classification Loss.} We use a cross-entropy loss between the predicted action probabilities and the corresponding generated pseudo-labels:
\begin{equation}
\label{eq:clsloss}
\mathcal{L}_{cls} = \frac{1}{\sum\nolimits_{n=1}^{N}(r_n-l_n)} \sum\limits_{n=1}^{N}\left(\sum\nolimits_{t=l_n}^{r_n-1} -\log {y}_t(c_{\tau_n})\right),
\end{equation}
where $y_t(c_{\tau_n})$ is the predicted probability that $x_t$ belongs to the $c_{\tau_n}$-th class, and the denominator of the former term, $\sum\nolimits_{n=1}^{N}(r_n-l_n)$, represents the total number of frames with pseudo-labels that contribute to the loss. It is worth noting that the frames within the ambiguous intervals have no pseudo-labels and are not calculated for cross-entropy loss.

\paragraph{Smoothing Loss.} We use the truncated mean squared error~\cite{farha2019ms} as the smoothing loss to improve over-segmentation errors:
\begin{equation}
\label{eq:smoloss} 
\mathcal{L}_{smo} = \frac{1}{TC} \sum\limits_{t = 1}^T \sum\limits_{c = 1}^C (max(\lvert \log {y}_t(c) - \log {y}_{t-1}(c) \rvert,\theta))^2,
\end{equation}
where $y_t(c)$ is the predicted probability that $x_t$ belongs to the $c$-th class, and \(\theta=4\) is a pre-set threshold. Besides, gradients are not calculated w.r.t. \({y}_{t-1}(c)\).

\paragraph{Confidence Loss.} The confidence loss~\cite{li2021temporal} is specifically designed for action segmentation under timestamp supervision:
\begin{equation}
\mathcal{L}_{conf} = \frac{1}{2(\tau_{N}-\tau_{1})} \sum\limits_{n=1}^{N} \left( \sum\nolimits_{t=\tau_{n-1}}^{\tau_{n+1}} \delta_{c_{\tau_n},t} \right),
\end{equation}
\begin{equation}
    \resizebox{.95\linewidth}{!}{$
            \displaystyle
\delta_{c_{\tau_n},t} = \begin{cases}
\max(0, \log y_{t}(c_{\tau_n}) - \log y_{t-1}(c_{\tau_n})) & if \enspace t \geq \tau_n,  \\
\max(0, \log y_{t-1}(c_{\tau_n}) - \log y_{t}(c_{\tau_n})) & if \enspace t < \tau_n, \\
\end{cases}$}
\end{equation}
where $2(\tau_{N}-\tau_{1})$ is the number of frames that contributed to the loss. The loss encourages the model to predict higher probabilities for low-confident intervals surrounded by high-confident intervals while suppressing those high-confident regions that are far from the labeled timestamp.

\paragraph{Clustering Loss.} To assist feature learning and clustering, we introduce a clustering loss commonly used in deep clustering, to distribute output features in the same action segment more compactly. IC aims to cluster similar features together and thus propagates pseudo-labels to frames within ambiguous intervals. As IC is similar to K-means clustering, we introduce the clustering loss~\cite{yang2017towards} commonly used in deep clustering into action segmentation for the first time. It can be formulated as:
\begin{equation}
\label{eq:cluloss} 
\mathcal{L}_{clu} = \frac{1}{\sum\nolimits_{n=1}^{N}(r_n-l_n)} \sum\limits_{n=1}^{N} \left(\sum\nolimits_{t=l_n}^{r_n-1} \Vert h_t - m_n \Vert_2^2\right),
\end{equation}
where $m_n=\frac{1}{r_n-l_n}\sum\nolimits_{t=l_n}^{r_n-1}h_t$ is the center vector of the $n$-th cluster. The loss encourages the high-level features of frames in the same cluster to be more compact so that the model is more discriminative for frames in different action segments. 
% In general, optimizing this loss alone will result in the model converging to a trivial solution, i.e., all features are mapped to the same point. Therefore, we optimize it together with the other three losses.

The overall loss function during the training phase is a weighted sum of the above four losses:
\begin{equation}
\label{eq:loss} 
\mathcal{L}^s = \mathcal{L}_{cls} + \lambda \mathcal{L}_{smo} + \beta \mathcal{L}_{conf} + \gamma \mathcal{L}_{clu},
\end{equation}
where $\lambda$, $\beta$ and $\gamma$ are hyper-parameters.
% to balance the contribution of each loss. 
% $\lambda$ is set to 0.15, $\beta$ is set to 0.075, and $\gamma$ is set to 0.15.

\section{Experiments}
\subsection{Datasets and Evaluation Metrics}
We empirically evaluate the performance of the proposed framwork on the following three public datasets. 
The \textbf{50Salads} dataset~\cite{50salads} contains 50 top-view videos with 17 action classes. On average, each video lasts for about 6.4 minutes and has about 20 action instances. The \textbf{GTEA} dataset~\cite{gtea} consists of 28 egocentric videos with 11 action classes. On average, there are about 20 action instances in each video. The \textbf{Breakfast} dataset~\cite{breakfast} contains 1,712 third-person view videos with 48 action classes. On average, each video is composed of about 6 action instances. We perform four-fold cross-validation on the GTEA and Breakfast datasets, and five-fold cross-validation on the 50Salads dataset. For evaluation, we report the average of all the splits.

We also use the segmental F1 score at overlapping thresholds 10\%, 25\%, 50\% (F1@{10,25,50}), segmental edit distance (Edit) measuring the difference between predicted segment paths and ground-truth segment paths, and frame-wise accuracy (Acc) as evaluation metrics.

\subsection{Implementation Details}

On all datasets, we represent each video as a sequence of visual features. We employ the I3D~\cite{i3d} features provided in~\cite{farha2019ms} as inputs. We utilize the same segmentation model following~\cite{li2021temporal}, which has two parallel stages for the first stage with kernel sizes 3 and 5 and passes the sum of both
outputs to the next stages. We train the model for 70 epochs with Adam optimizer~\cite{adam}. Similar to~\cite{li2021temporal}, in the first 50 epochs, we only use the annotated timestamps to train to find a good initialization. Subsequently, we apply the IC algorithm to train for 20 epochs. We use the learning rate 0.0001 for Breakfast and 0.0005 for GTEA and 50Salads. We set $\lambda$, $\beta$, $\gamma$ to 0.15, 0.075 and 0.15 respectively.

\subsection{Comparison with the State-of-the-art}

\begin{table}[t]
    \centering
    \resizebox{0.98\columnwidth}{!}{
    \begin{tabular}{c|l|ccccc}
    \toprule
    Supervision & Method & \multicolumn{3}{c}{F1@\{10,25,50\}} & Edit  & Acc \\
    \midrule\midrule
    
    \multirow{4}{*}{Full}
        & MS-TCN  & 76.3  & 74.0  & 64.5  & 67.9 & 80.7   \\
        & MS-TCN++  & 80.7  & 78.5  & 70.1  & 74.3 & 83.7   \\
        & BCN  & 82.3  & 81.3  & 74.0  & 74.3 & 84.4   \\
        & ASRF  & 84.9  & 83.5  & 77.3  & 79.3 & 84.5   \\

    \midrule
    
    \multirow{2}{*}{Transcript}
        & NN-Viterbi  & -  & -  & -  & - & 49.4  \\ 
        & CDFL  & -  & -  & - & -  & 54.7  \\
        
    \midrule

    \multirow{3}{*}{Timestamp}
        & Li \textit{et al.}  & 73.9  & 70.9  & 60.1  & 66.8 & 75.6  \\
        & GCN & 75.1  & 72.3  & 61.0  & 67.6 & 75.1  \\
        & Ours  & \textbf{77.3} & \textbf{74.7} & \textbf{63.7} & \textbf{70.1} & \textbf{78.6}\\

    \bottomrule
    \end{tabular}}
    \caption{Comparison with the state-of-the-art on 50Salads.}
    \label{tab:sota_50salads}
\end{table}
\begin{table}[t]
    \centering
    \resizebox{0.98\columnwidth}{!}{
    \begin{tabular}{c|l|ccccc}
    \toprule
    Supervision & Method & \multicolumn{3}{c}{F1@\{10,25,50\}} & Edit  & Acc \\
    \midrule\midrule
    
    \multirow{4}{*}{Full}
        & MS-TCN  & 85.8  & 83.4  & 69.8  & 79.0 & 76.3   \\
        & MS-TCN++  & 88.8  & 85.7  & 76.0  & 83.5 & 80.1   \\
        & BCN  & 88.5  & 87.1  & 77.3  & 84.4 & 79.8   \\
        & ASRF  & 89.4  & 87.8  & 79.8  & 83.7 & 77.3   \\

    \midrule

    \multirow{3}{*}{Timestamp}
        & Li \textit{et al.}  & 78.9  & 73.0  & 55.4  & 72.3 & 66.4  \\
        & GCN  & 81.5  & 77.5  & 60.8  & 75.6 & 66.1  \\
        & Ours  & \textbf{83.7} & \textbf{79.8} & \textbf{65.4} & \textbf{77.2} & \textbf{70.1}\\

    \bottomrule
    \end{tabular}}
    \caption{Comparison with the state-of-the-art on GTEA.}
    \label{tab:sota_gtea}
\end{table}
\begin{table}[!t]
    \centering
    \resizebox{0.98\columnwidth}{!}{
    \begin{tabular}{c|l|ccccc}
    \toprule
    Supervision & Method & \multicolumn{3}{c}{F1@\{10,25,50\}} & Edit  & Acc \\
    \midrule\midrule
    
    \multirow{4}{*}{Full}
        & MS-TCN & 52.6  & 48.1  & 37.9  & 61.7 & 66.3   \\
        & MS-TCN++ & 64.1  & 58.6  & 45.9  & 65.6 & 67.6   \\
        & BCN & 68.7  & 65.5  & 55.0  & 66.2 & 70.4   \\
        & ASRF & 74.3  & 68.9  & 56.1  & 72.4 & 67.6   \\

    \midrule
    
    \multirow{2}{*}{Transcript}
        & NN-Viterbi & -  & -  & -  & - & 43.0  \\ 
        & CDFL & -  & -  & - & -  & 50.2  \\
        
    \midrule

    \multirow{2}{*}{Set}
        & SCT & -  & -  & -  & - & 30.4  \\ 
        & ACV & -  & -  & - & -  & 33.4  \\
        
    \midrule

    \multirow{3}{*}{Timestamp}
        & Li \textit{et al.}  & 70.5  & 63.6  & 47.4 & 69.9 & 64.1 \\
        & GCN  & 67.9 & 61.0 & 45.3 & 67.0 & 61.4 \\
        & Ours  &  \textbf{71.2}  & \textbf{64.6} & \textbf{48.9} & \textbf{71.6} & \textbf{65.7} \\

    \bottomrule
    \end{tabular}}
    \caption{Comparison with the state-of-the-art on Breakfast.}
    \label{tab:sota_breakfast}
\end{table}

We compare our method with recent state-of-the-art approaches, including four fully supervised methods (MS-TCN~\cite{farha2019ms}, MS-TCN++~\cite{li2020ms}, BCN~\cite{wang2020bcn}, ASRF~\cite{asrf}), two transcript-supervised methods (NN-Viterbi~\cite{nnviterbi}, CDFL~\cite{cdfl}), two set-supervised methods (SCT~\cite{sct}, ACV~\cite{acv}),  and two timestamp-supervised methods (Li et al.~\shortcite{li2021temporal}, GCN~\cite{khan2022timestamp}). 

Results for three datasets are shown in Table~\ref{tab:sota_50salads},~\ref{tab:sota_gtea},~\ref{tab:sota_breakfast}. The best results are highlighted in bold. Under timestamp supervision, our method achieves state-of-the-art on all datasets. Besides, we find that our timestamp-supervised approach significantly outperforms all the set-supervised and transcript-supervised methods by a large margin. Furthermore, our method is comparable to or even better than some fully supervised models such as MS-TCN on the 50Salads and Breakfast datasets. In particular, on the largest Breakfast dataset, the performance is even comparable to one of the most advanced fully supervised methods, BCN.

% For example, on the 50Salads dataset, our method gives \textbf{2.2}\% (75.1$\to$77.3), \textbf{2.4}\% (72.3$\to$74.7), \textbf{2.7}\% (61.0$\to$63.7) improvements on three F1 scores, \textbf{2.5}\% (67.6$\to$70.1) improvement on the segmental edit distance, and \textbf{3.0}\% (75.6$\to$78.6) improvement on the accuracy. 

\subsection{Ablation Study}

\begin{table}[ht]
    \centering
    \resizebox{1\columnwidth}{!}{
    \setlength{\tabcolsep}{0.9mm}{
    \begin{tabular}{ccc|cc|cc|cc}
    \toprule
    Energy  & Cons & Temp & \multicolumn{2}{c|}{GTEA} & \multicolumn{2}{c|}{50Salads} & \multicolumn{2}{c}{Breakfast}\\
    Func & K-m & AGNES & Acc   & Label & Acc   & Label & Acc   & Label \\
    \midrule\midrule
    
    \ding{51} &            &            & 75.6  & 100   & 79.9 & 100   & 72.7  & 100 \\
               & \ding{51} &            & 77.1  & 100   & 77.9 & 100   & 75.2  & 100 \\
               &            & \ding{51} & 78.4  & 100   & 81.5 & 100   & 75.1  & 100 \\
    % \ding{51} & \ding{51} &            & 80.2  & 87.3  & 86.5 & 79.1  & 81.0  & 77.1 \\
    % \ding{51} &            & \ding{51} & 82.9  & 82.1  & 88.4 & 80.0  & 80.4  & 78.5 \\
    %            & \ding{51} & \ding{51} & 82.9  & 84.3  & 86.1 & 82.3  & 81.3 & 80.5 \\
    \ding{51} & \ding{51} & \ding{51} & 84.0 & 76.9 & 90.0 & 70.7  & 84.2 & 68.1 \\
    
    \midrule
    \ding{51}$^*$ & \ding{51}$^*$ & \ding{51}$^*$ & \textbf{86.8} & 60.4  & \textbf{95.5} & 51.2 & \textbf{88.8} & 50.9 \\

    \bottomrule
    \end{tabular}
    }}
    \caption{Comparison of the quality of pseudo-labels generated by different methods. \textit{Energy Func}, \textit{Cons K-m}, and \textit{Temp AGNES} are abbreviations of the energy function, constrained k-medoids, and temporal AGNES, respectively. \textit{Acc} and \textit{Label} denote the label accuracy and label rate, respectively.}
    \label{tab:abla_ple}
\end{table}

\begin{figure}[ht]
\begin{center}
\includegraphics[width=0.82\columnwidth]{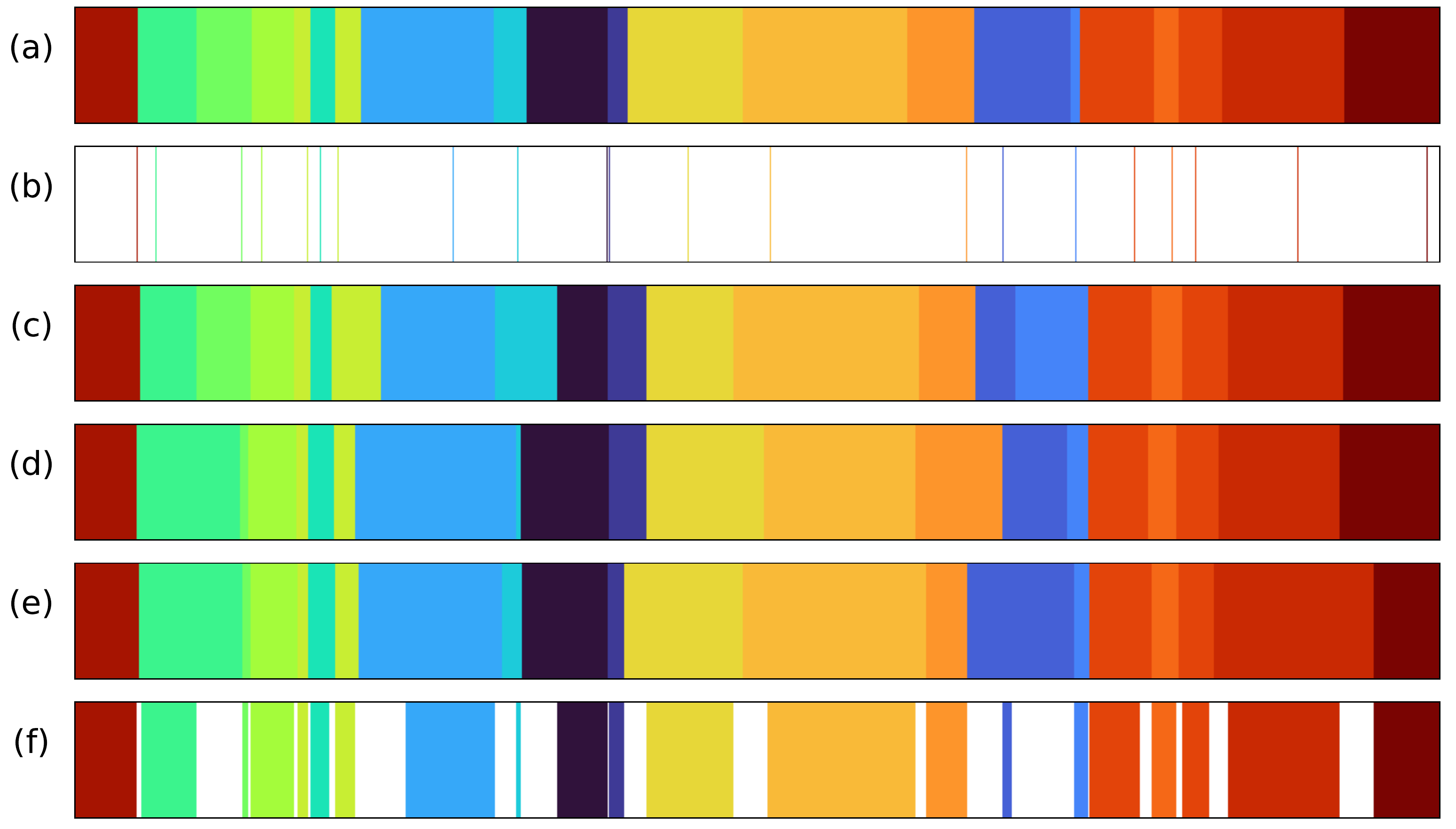}
\caption{Visualization of pseudo-label sequences generated by several methods: (a) ground truth, (b) timestamp supervision, (c) energy function, (d) constrained k-medoids, (e) temporal AGNES, (f) pseudo-label ensembling. Frames in white intervals have no labels.}
\label{fig:ple}
\end{center}
\vspace{-0.2 cm}
\end{figure}

\paragraph{Impact of Pseudo-Label Ensembling.} Given timestamp supervision and frame-wise visual features, we generate pseudo-labels with different methods and evaluate them with label accuracy and label rate in Table~\ref{tab:abla_ple}.  We find that our temporal AGNES algorithm outperforms the other two. In the penultimate row, we can obtain pseudo-labels with both a high label rate and high accuracy when we ensembling these three algorithms, i.e., PLE. In practice, we further divide the visual features into RGB features and optical flow features, and ensemble these two types of features separately to get the final results, which are listed in the last row. The final pseudo-labels achieve the highest accuracy and over 50\% label rate. We visualize the generated pseudo-label sequences in Figure~\ref{fig:ple}. The provided timestamp supervision signal is too sparse because the label rate is only $0.4\%$. We can generate full pseudo-label sequences by any clustering algorithm, but there are too many wrong labels. PLE makes a trade-off between the label rate and accuracy.

\begin{table}
    \centering
    \begin{tabular}{lccccc}
    \toprule
    Method & \multicolumn{3}{c}{F1@\{10,25,50\}} & Edit  & Acc \\
    \midrule\midrule
    
    PLE & 72.2 & 69.0 & 57.3 & 64.4 & 76.6 \\
    Temp AGNES & 72.2 & 68.3 & 56.3 & 64.5 & 73.2 \\
    \midrule
    Li \textit{et al.} & 75.4  & 72.4  & 61.8  & 68.1  & 76.8 \\
    IC (Li \textit{et al.})  & 74.7 & 71.8 & 62.1 & 66.2 & 77.6 \\
    \midrule
    IC (Ours) & \textbf{76.0} & \textbf{73.2} & \textbf{63.7} & \textbf{68.5} & \textbf{78.5}\\
    
    \bottomrule
    \end{tabular}
    \caption{Comparison of supervision generated by different methods. The first two and the middle two rows represent that pseudo-labels are fixed and updated during training, respectively. \textit{IC (Li et al.)} refers to using the energy function to generate pseudo-labels for the ambiguous intervals in pseudo-label sequences during IC. \textit{IC (Ours)} is IC we propose.}
    \label{tab:abla_IC}
\end{table}

\begin{figure}[ht]
\begin{center}
\includegraphics[width=0.8\columnwidth]{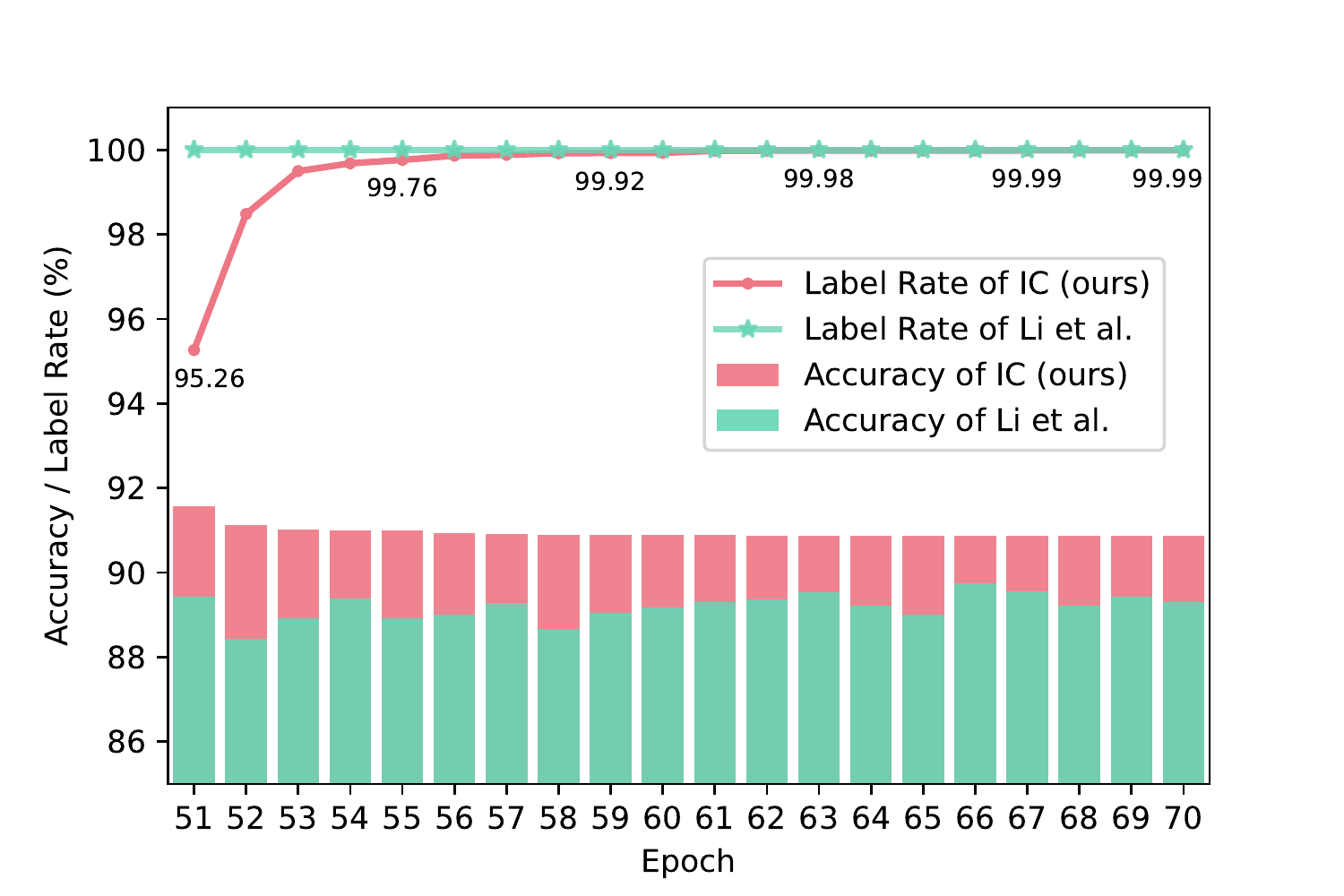}
\caption{Comparison of the quality of the pseudo-labels generated by our proposed IC and the method of Li et al. during training. The method of Li et al. does not consider ambiguous intervals, and thus the label rate is always 100\%.}
\label{fig:acc_label_rate}
\end{center}
\vspace{-0.3 cm}
\end{figure}

\paragraph{Ablations about Iterative Clustering.} Iterative clustering (IC) is an important part of our framework for which we conduct detailed ablation studies.  We use pseudo-label sequences updated by IC to train the model. We can also replace IC with other methods to generate pseudo-labels for training. In Table~\ref{tab:abla_IC}, we first generate pseudo-label sequences with and without ambiguous intervals by PLE and Temporal AGNES to train the model directly. The pseudo-label sequences depend only on the visual features and do not change during training. Therefore, these two methods have the worst performance. The methods in the last three rows of the table all iteratively update the pseudo-labels based on high-level features during training. Li et al.~\shortcite{li2021temporal} use the energy function to generate full pseudo-label sequences. The modified version of IC, called IC (Li et al.), uses the energy function to generate pseudo-labels for the frames within ambiguous intervals, which also yields full pseudo-label sequences. IC (Ours) generates pseudo-label sequences with ambiguous intervals to train the model and achieves the best results, which illustrates the importance of ambiguous intervals. As the training epoch increases, Figure~\ref{fig:acc_label_rate} shows the variation of accuracy and label rate of pseudo-labels generated by energy function and IC. The accuracy of IC is always higher than that of Li et al., and the label rate of IC is increasing, indicating that the range of ambiguous intervals is decreasing.

\begin{figure}[ht]
\begin{center}
\includegraphics[width=0.76\columnwidth]{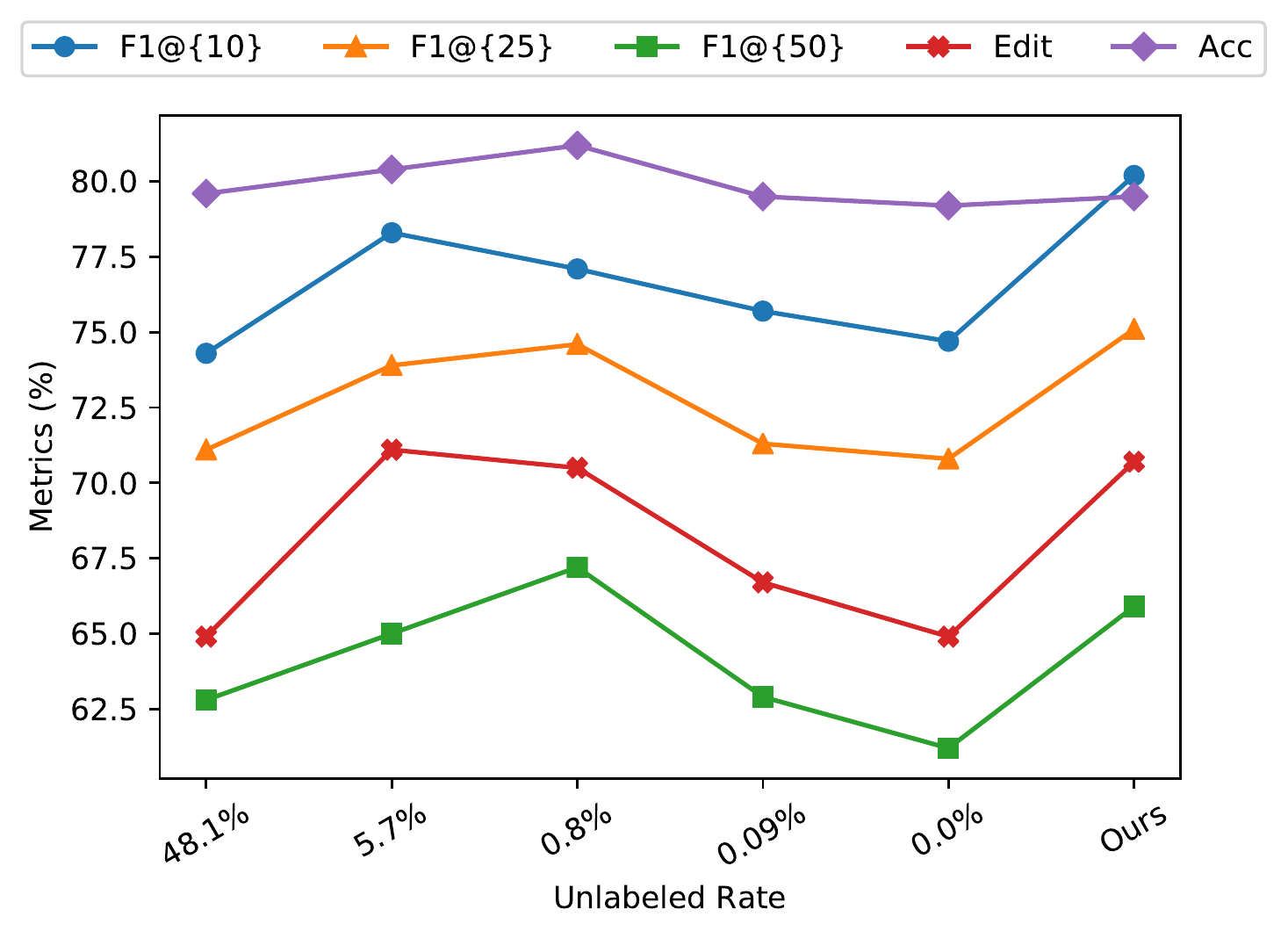}
\caption{Comparison of the performance of training the model using ground-truth label sequences with different unlabeled rates, which can be obtained by removing the labels from the ground truth based on ambiguous intervals generated by the IC. For comparison, the last column shows the performance of our IC.}
\label{fig:unlabelrate}
\end{center}
\end{figure}

\paragraph{Effect of ambiguous intervals.} To further explore the effect of ambiguous intervals, we remove the labels of ambiguous intervals from these ground-truth label sequences under full supervision. Then we use the resulting label sequences with a certain unlabeled rate to train the model. Since the range of ambiguous intervals is decreasing during IC, we can record multiple sets of ambiguous intervals, which correspond to multiple label sequences with different unlabeled rates. As shown in Figure~\ref{fig:unlabelrate}, the frames within the ambiguous intervals generated by PLE take up 48.1\% of the total frames. So we can use the label sequences with an unlabeled rate of 48.1\% to train the model. Similarly, the unlabeled rates of 5.7\%, 0.8\% and 0.09\% represent the ambiguous intervals when IC runs to three intermediate epochs. The unlabeled rate of 0\% represents full supervision. The model performs best when the unlabeled rate is 0.8\%, which indicates that about 0.8\% labels in ground-truth label sequences are ambiguous and useless. We should train the model without these labels instead of using all the labels. Our IC does not use the information in the ground truth, but still achieves comparable results to the setting of unlabeled rate 0.8\% and outperforms the fully supervised setting.

\begin{table}
    \centering
    \resizebox{0.95\columnwidth}{!}{
    \begin{tabular}{l|c|ccccc}
    \toprule
    Dataset & $\mathcal{L}_{clu}$ & \multicolumn{3}{c}{F1@\{10,25,50\}} & Edit  & \multicolumn{1}{c}{Acc} \\
    \midrule\midrule

    \multirow{2}{*}{GTEA} 
    & \ding{55} & 82.0 & 79.1 & 64.5 & 76.1 & 69.5\\
    & \ding{51} & \textbf{83.7} & \textbf{79.8} & \textbf{65.4} & \textbf{77.2} & \textbf{70.1}\\
    
    \midrule    

    \multirow{2}{*}{50Salads} 
    & \ding{55} & 76.0 & 73.2 & \textbf{63.7} & 68.5 & 78.5\\
    & \ding{51} & \textbf{77.3} & \textbf{74.7} & \textbf{63.7} & \textbf{70.1} & \textbf{78.6}\\

    \bottomrule
    \end{tabular}}
    \caption{Impact of the clustering loss $\mathcal{L}_{clu}$ on two datasets.}
    \label{tab:abla_cluloss}
\end{table}

\begin{figure}[t]
\begin{center}
\includegraphics[width=0.84\columnwidth]{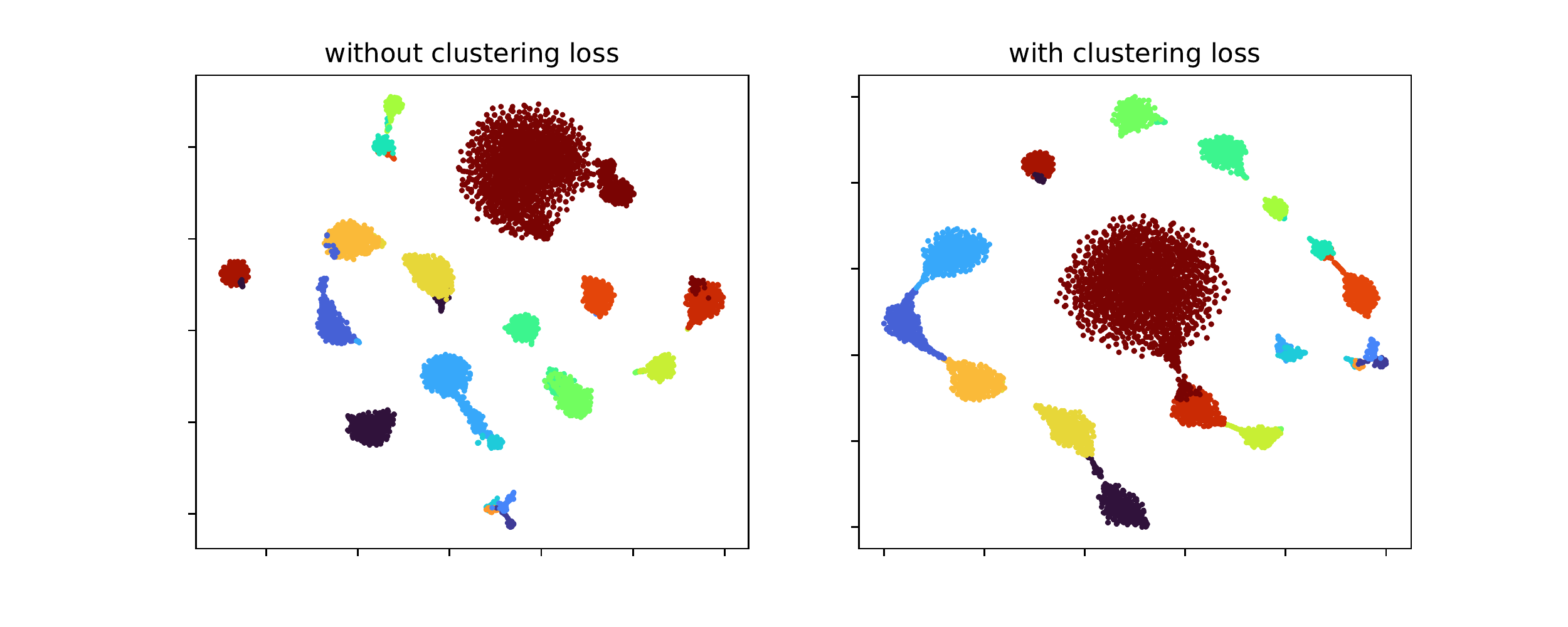}
\caption{With or without the clustering loss, T-SNE visualization of all the learned output features in one video in the 50Salads dataset.}
\label{fig:tsne}
\end{center}
\vspace{-0.3 cm}
\end{figure}

\paragraph{Impact of the Clustering Loss.} To verify the effectiveness of our proposed clustering loss, we train the segmentation model with and without it. The metrics for both settings on two datasets are presented in Table~\ref{tab:abla_cluloss}. Additionally, we show the TSNE visualization of output features in Figure~\ref{fig:tsne}. We observe that the model's output features become more discriminative with the help of clustering loss, leading to improvements in all metrics.

\section{Conclusion}

From the perspective of clustering, we propose pseudo-label ensembling and iterative clustering and introduce clustering loss to compose a framework for timestamp-supervised action segmentation. For the first time, we consider the ambiguity of action segment boundaries. Instead of directly generating pseudo-labels for all frames, our approach generates labels only for semantically explicit frames, which avoids the negative impact of inconsistent labels for ambiguous intervals at action segment boundaries. Experiments verify the effectiveness of our framework.

\appendix
\section{Temporal AGNES}
\label{sec:agnes}

Algorithm~\ref{alg:AGNES} summarizes the steps of our clustering algorithm, which we call the temporal AGNES algorithm. Function $\textup{dist}(\cdot,\cdot)$ calculates the Euclidean distance between visual features of two frames, and matrix $D$ contains all the distances between any pair of frames. In contrast to the original AGNES algorithm, we make the following three changes. First, different timestamps should have distinct labels and cannot be merged. Therefore, we set the distance between different timestamp frame features to infinity to prevent them from being merged (step 9). Second, we only merge temporally adjacent clusters. Therefore, we only record and delete the boundaries of segments when merging two temporally contiguous clusters (step 3 and step 15). Third, since each action segment in the video contains a single timestamp, the algorithm stops when the number of clusters equals the number of timestamps (step 11). We utilize average linkage as the linkage criterion when merging clusters, i.e., the distance between a pair of clusters is determined by the average distance between frames (step 13 and step 14).

\begin{algorithm}[h]
    \small
    \caption{Temporal AGNES clustering algorithm}
    \label{alg:AGNES}
    \renewcommand{\algorithmicrequire}{\textbf{Input:}}
    \renewcommand{\algorithmicensure}{\textbf{Output:}}
    \begin{algorithmic}[1]
        \REQUIRE visual features: $[x_1,x_2,...,x_T]$, timestamp supervision: $\{c_{\tau_1},c_{\tau_2},...,c_{\tau_N}\}$
        \ENSURE pseudo-label sequence: $[\hat{c}_1,\hat{c}_2,...,\hat{c}_T]$
        
        \STATE $q \gets T$ 
        % \Comment{初始化聚类簇数}
        \FOR{$i=1,2,...,T$}
            \STATE $b_i, d_i \gets i, \textup{dist}(x_i, x_{i+1})$ 
            % \Comment{初始化簇的边界和簇间距离}
            \FOR{$j=1,2,...,T$}
                \STATE $D[i,j] = \textup{dist}(x_i, x_j)$ 
                % \Comment{初始化特征距离矩阵$D$}
            \ENDFOR
        \ENDFOR
        
        \FOR{$n=1,2,...,N-1$}
            \STATE $D[\tau_{n},\tau_{n+1}] \gets \infty $
            % \Comment{防止属于不同动作段的时间戳被聚为一类}
        \ENDFOR

         \WHILE{$q > N$}
            \STATE $i = \mathop{\mathrm{argmin}}_{j}{d_j}$
            % \Comment{find minimum cluster distance}
            \STATE $d_{i-1} \gets \frac{d_{i-1}*(b_{i}-b_{i-1})*(b_{i+1}-b_{i}) + \sum\nolimits_{j=b_{i-1}}^{b_i}\sum\nolimits_{k=b_{i+1}}^{b_{i+2}} D[j,k]}{(b_{i}-b_{i-1})*(b_{i+2}-b_{i})}$ 
            % \Comment{更新簇间距离}
            \STATE $d_{i} \gets \frac{d_{i}*(b_{i+2}-b_{i+1})*(b_{i+3}-b_{i+2}) + \sum\nolimits_{j=b_{i}}^{b_{i+1}}\sum\nolimits_{k=b_{i+2}}^{b_{i+3}} D[j,k]}{(b_{i+3}-b_{i+2})*(b_{i+2}-b_{i})}$ 
            % \Comment{更新簇间距离}
            \STATE \textbf{delete} $b_{i+1}, d_{i+1}$           
            % \Comment{合并两个簇} 
            \STATE $q \gets q-1$
            % \Comment{更新聚类簇数}
        \ENDWHILE  

        \FOR{$n=1,2,...,N$}
        % \Comment{根据簇的边界生成伪标签序列}
        \STATE $[\hat{c}_{b_n},\hat{c}_{b_{n}+1},...,\hat{c}_{b_{n+1}-1}] \gets [c_{\tau_n},c_{\tau_n},...,c_{\tau_n}]$ 
        % \Comment{属于同一个动作段}
        \ENDFOR
        
    \end{algorithmic}
\end{algorithm}

% \section*{Acknowledgments}

% The preparation of these instructions and the \LaTeX{} and Bib\TeX{}
% files that implement them was supported by Schlumberger Palo Alto
% Research, AT\&T Bell Laboratories, and Morgan Kaufmann Publishers.
% Preparation of the Microsoft Word file was supported by IJCAI.  An
% early version of this document was created by Shirley Jowell and Peter
% F. Patel-Schneider.  It was subsequently modified by Jennifer
% Ballentine, Thomas Dean, Bernhard Nebel, Daniel Pagenstecher,
% Kurt Steinkraus, Toby Walsh, Carles Sierra, Marc Pujol-Gonzalez,
% Francisco Cruz-Mencia and Edith Elkind.

%% The file named.bst is a bibliography style file for BibTeX 0.99c
\bibliographystyle{named}
\bibliography{ijcai23}

\end{document}